%% file: arxiv_main.tex
\documentclass[a4paper]{article}

\usepackage{chngcntr}
\usepackage[utf8x]{inputenc}
\usepackage[T1]{fontenc}

\usepackage[round]{natbib}

\usepackage[papersize={8.5in,11in},top=1in,bottom=0.75in,left=0.75in,right=0.75in]{geometry}
\usepackage[colorinlistoftodos]{todonotes}
\usepackage[colorlinks=true, allcolors=blue]{hyperref}
\usepackage{authblk}
\usepackage{enumitem}
\usepackage{algorithm}
\usepackage{algorithmic}
\input{commands.tex}

\title{Optimization and Generalization of Regularization-Based Continual Learning: a Loss Approximation Viewpoint}

\author[1]{Dong Yin \thanks{dongyin@google.com}}
\author[1]{Mehrdad Farajtabar \thanks{farajtabar@google.com}}
\author[1]{Ang Li \thanks{anglili@google.com}}
\author[1]{Nir Levine \thanks{nirlevine@google.com}}
\author[1]{Alex Mott \thanks{alexmott@google.com}}
\affil[1]{DeepMind, Mountain View}

\begin{document}
\maketitle

\begin{abstract}
Neural networks have achieved remarkable success in many cognitive tasks. However, when they are trained sequentially on multiple tasks without access to old data, their performance on early tasks tend to drop significantly.
This problem is often referred to as \emph{catastrophic forgetting}, a key challenge in \emph{continual learning} of neural networks.
The regularization-based approach is one of the primary classes of methods to alleviate catastrophic forgetting.
In this paper, we provide a novel viewpoint of regularization-based continual learning by formulating it as a second-order Taylor approximation of the loss function of each task.
This viewpoint leads to a unified framework that can be instantiated to derive many existing algorithms such as Elastic Weight Consolidation and Kronecker factored Laplace approximation.
Based on this viewpoint, we study the optimization aspects (\ie convergence) as well as generalization properties (\ie finite-sample guarantees) of regularization-based continual learning. Our theoretical results indicate the importance of accurate approximation of the Hessian matrix.
The experimental results on several benchmarks provide an empirical validation of our theoretical findings.
\end{abstract}

\section{Introduction}\label{sec:intro}
Neural networks are achieving human-level performance on many cognitive tasks including image classification~\citep{krizhevsky2012imagenet} and speech recognition~\citep{hinton2006fast}. However, as opposed to humans, their acquired knowledge is comparably volatile and can be easily dismissed. Especially, the \emph{catastrophic forgetting} phenomenon refers to the case when a neural network forgets the past tasks if it is not allowed to retrain or reiterate on them again~\citep{goodfellow2013empirical,mccloskey1989catastrophic}.

\emph{Continual learning} is a research direction that deals with the catastrophic forgetting problem. Recent works tried to tackle this issue from a variety of perspectives. Most approaches can be categorized into three classes, \ie, the regularization-based~ \citep{kirkpatrick2017overcoming,zenke2017continual,ritter2018online,farajtabar2019orthogonal}, expansion-based~ \citep{rusu2016progressive,yoon2018lifelong}, and replay-based methods~ \citep{lopez2017gradient,chaudhry2018efficient}.

Although various algorithms have been proposed, few efforts were spent on the theoretical foundations of continual learning. Take the regularization-based approach as an example; in these algorithms, we usually construct a quadratic regularization function which penalizes the updates to the weights that are important to prior tasks. In most works, the regularization functions are designed based on heuristics without a theoretically rigorous understanding of how they affect the optimization and generalization of the learning algorithm, and how they mitigate catastrophic forgetting. We believe a solid theoretical foundation can enable better design and development of efficient continual learning algorithms.

In this paper, we tackle the optimization and generalization aspects of regularization-based continual learning. We study a unified framework for these algorithms from the perspective of loss landscapes. More specifically, in our framework at the end of each task we approximate the loss function by estimating its second-order Taylor expansion. The approximation is used as a surrogate added to the loss function of the future tasks. We show that many existing algorithms such as Elastic Weight Consolidation (EWC)~\citep{kirkpatrick2017overcoming} and Kronecker factored Laplace approximation~\citep{ritter2018online} are special cases of our framework.

More importantly, equipped with this loss function approximation viewpoint, we are able to conduct rigorous analysis of the regularization-based algorithms. In summary, we make the following theoretical contributions:

\begin{list}{\labelitemi}{\leftmargin=1.5em}

\item To study the optimization and generalization properties of regularization-based continual learning, we provide a loss approximation framework that unifies a few important continual learning methods.

\item We leverage the unified framework to prove a \emph{sufficient and worst-case necessary} condition under which by conducting gradient descent on the approximate loss function, we can still minimize the actual loss function. We identify the accuracy of Hessian matrix approximation, the moving distance in parameter space between tasks, and high order derivatives as important factors in the optimization procedure of the algorithms.

\item We further provide convergence analysis of regularization-based continual learning for both non-convex and convex loss functions.

\item We also investigate the generalization property of the algorithms under the loss approximation framework and provide finite-sample guarantees.
\end{list}

In addition to the theoretical contributions, we also present experimental evidence on three common continual learning benchmarks to validate our theoretical findings using the loss approximation viewpoint.

\paragraph*{Notation} Throughout this paper, for any positive integer $N$, we define $[N] :=\{1,2,\ldots, N\}$. We use $\twonms{\cdot}$ to denote the $\ell_2$ norm of vectors and the operator norm of matrices.

\section{Related Work}\label{sec:related_work}

\paragraph*{Continual learning algorithms}
Generally speaking, three classes of algorithms exist to overcome catastrophic forgetting: \emph{regularization-based}, \emph{replay-based}, and \emph{expansion-based} methods.

The \emph{regularization-based} approaches are the main focus of this paper. These methods usually impose limiting constraints on the weight updates of the neural network according to some relevance score for previous knowledge~\citep{kirkpatrick2017overcoming,zenke2017continual,ritter2018online, mirzadeh2020dropout,farajtabar2019orthogonal}. They can also be combined with other methods~\citep{nguyen2017variational,titsias2019functional}. A large fraction of regularization-based algorithms construct a quadratic regularization function at the end of each task and combine it with the loss of future tasks~\citep{kirkpatrick2017overcoming,zenke2017continual,ritter2018online,aljundi2018memory,park2019continual}. In this paper, we show that our loss function approximation viewpoint provides a general framework for understanding these methods.

The \emph{expansion-based} methods allocate new neurons or layers or modules to accommodate new tasks while utilizing the shared representation learned from previous ones~\citep{rusu2016progressive, xiao2014error,yoon2018lifelong,li2019learn,Jerfel2018ReconcilingMA}. The \emph{replay-based} methods store previous data or, alternatively, train a generative model of them and replay samples from them interleaved with samples drawn from the current task~\citep{shin2017continual,kamra2017deep,zhang2019prototype,rios2018closed,luders2016continual,lopez2017gradient}.

\paragraph*{Theoretical understanding for continual learning}
One prior work that uses second-order Taylor expansion to approximate the loss functions of previous tasks is the Efficient Lifelong Learning Algorithm (ELLA)~\citep{ruvolo2013ella}. We emphasize that ELLA is actually a special case in our framework. ELLA aims to construct a latent dictionary such that the model parameters for each task can be represented as a sparse linear combination of the items in the dictionary. Using the terminology of neural networks, this is equivalent to a one hidden layer linear network (\ie the activation function is identity) with multi-head structure, where each head corresponds to each task. In contrast, our framework does not rely on specific neural network architectures and thus can be applied to more general problems. As for the theoretical guarantees, \cite{ruvolo2013ella} only showed the convergence of the \emph{approximate loss}, whereas in this paper, we focus on mitigating catastrophic forgetting, \ie the gap between the approximate loss and the actual loss and how this gap affects the optimization and generalization of the learning problem.

Another line of theoretical research focuses on regret analysis for continual learning~\citep{alquier2017regret,finn2019online,wu2019lifelong}. These studies aim to analyze the behavior of the algorithm \emph{during the course of training}. In comparison, we focus on the forgetting problem, \ie the performance of the \emph{final model}. A recent work~\citep{bennani2020generalisation} studies the generalization property of the OGD algorithm~\citep{farajtabar2019orthogonal}, which does not use the quadratic regularization loss. Another important but different topic in continual/multi-task learning is the generalization to unseen tasks~\citep{baxter1998theoretical,maurer2016benefit,pentina2015multi}. A few other works try to obtain a better understanding of continual learning via empirical methods~\citep{nguyen2019toward,farquhar2019unifying,mirzadeh2020understanding}.

\section{Problem Formulation}
In this section, we present the continual learning formulation studied in this paper.
We consider a sequence of $K$ supervised learning tasks $\T_k$, $k\in[K]$. For task $\T_k$, there is an unknown distribution $\D_k$ over the space of feature-label pairs $\X\times\Y$. Let $\W\subseteq\R^d$ be a model parameter space, and for the $k$-th task, let $\ell_k(w; x, y):\W \mapsto \R$ be the loss function of $w$ associated with data point $(x, y)$. The population loss function of task $\T_k$ is defined as
\[
L_k(w) := \EE_{(x, y)\sim \D_k} \ell_k(w; x, y).
\]
Our general objective is to learn a parametric model with minimized population loss over all the $K$ tasks. More specifically, in continual learning the learner follows the following protocol: When learning on the $k$-th task, the learner obtains access to $n_k$ data points $(x_{k,i}, y_{k,i})$, $i\in[n_k]$ sampled i.i.d. according to $\D_k$ and we define
\[
\widehat{L}_k(w) := \frac{1}{n_k}\sum_{i=1}^{n_k} \ell_k(w; x_{k,i}, y_{k,i})
\]
as the empirical loss function; the learner then updates the model parameter $w$ using these $n_k$ training data, and after the training procedure is finished, the learner loses access to them, but can store some side information about the task. The goal is to avoid forgetting previous tasks when trained on new tasks by utilizing the side information.

\section{Regularization-Based Approaches and Loss Approximation Viewpoint}\label{sec:approach}

As mentioned in Section~\ref{sec:related_work}, one important approach to solving continual learning problems is to use regularization. In these algorithms, at the end of each task, we construct a regularization function of the model parameters, usually quadratic, and when training future tasks, we minimize the regularized loss function. In this section, we derive a unified framework for regularization-based continual learning through the lens of loss function approximation, and demonstrate that many existing algorithms can be considered as variants of our framework.

To measure the effectiveness of a continual learning algorithm, we use a simple criterion that after each task, we want the average population loss over all the tasks 
to be small. This means that, for every $k\in[K]$, after training on $\T_k$, we want to find $w$ to minimize $\frac{1}{k}\sum_{k'=1}^k L_{k'}(w)$. Since optimizing the loss function is the key to training a good model, we begin with a straightforward method for continual learning: storing the second-order Taylor expansion of the empirical loss function, and using it as an approximation of the loss function for an old task when training on new tasks.
We start with a simple setting. Suppose that there are two tasks, and at the end of $\T_1$, we obtain a model $\widehat{w}_1$. Then we compute the gradient and Hessian matrix of $\hatL_1(w)$ at $\hatw_1$, and construct the second-order Taylor expansion of $\hatL_1(w)$ at $\hatw_1$:
\begin{align*}
L_1^{\text{prox}}(w) = \hatL_1(\hatw_1) + (w - \hatw_1)^\top\nabla \hatL_1(\hatw_1) + \frac{1}{2}(w - \hatw_1)^\top \nabla^2 \hatL_1(\hatw_1) (w - \hatw_1).
\end{align*}
When training on $\T_2$, we try to minimize $\frac{1}{2}(L_1^{\text{prox}}(w) + \hatL_2(w))$. The basic idea of this design is that, we hope in a neighborhood around $\hatw_1$, the quadratic function $L_1^{\text{prox}}(w)$ stays as a good approximation of $\hatL_1(w)$, and thus approximately still minimizes the average of the empirical loss functions $\frac{1}{2}(\hatL_1(w) + \hatL_2(w))$, which in the limit generalizes to the population loss function $\frac{1}{2}(L_1(w) + L_2(w))$.

Formally, let $\hatw_k$ be the model that we obtain at the end of the $k$-th task. We define the approximation of the sum of the first $k$ empirical loss functions as
\begin{equation}\label{eq:loss_prox}
L_k^{\text{prox}}(w) = \sum_{k'=1}^k \underbrace{\hatL_{k'}(\hatw_{k'})}_{\text{(a)}} + \underbrace{(w - \hatw_{k'})^\top\nabla \hatL_{k'}(\hatw_{k'})}_{\text{(b)}}
+ \underbrace{\frac{1}{2}(w - \hatw_{k'})^\top H_{k'} (w - \hatw_{k'})}_{\text{(c)}},
\end{equation}
where, $H_{k'}$ denotes the Hessian matrix $\nabla^2 \hatL_{k'}(\hatw_{k'})$ or its approximation.
We construct $L_k^{\text{prox}}(w)$ at the end of task $k$, and when training on task $k+1$, we minimize $\frac{1}{k+1}(L_k^{\text{prox}}(w) + \hatL_{k+1}(w))$.

One important assumption that we make throughout this work is the validity of the second-order Taylor expansion as an approximation to the loss functions. While this might seem as a crude approximation in general, we argue this has merit for our setup. More specifically, we rely on the wealth of studies on overparameterized models where the loss tends to be very well behaved and almost convex in a reasonable neighborhood around the minima~\citep{saxe2013exact,choromanska2015loss,goodfellow2014qualitatively}. Moreover, there is a similar observation in the recently proposed NTK regime~
\citep{jacot2018neural}.
In continual learning, similar assumptions are made by most approaches that rely on approximating the posterior on the weights by a Gaussian~\citep{kirkpatrick2017overcoming} or a first-order approximation of the loss surface around the optimum~\citep{farajtabar2019orthogonal} or an explicit Taylor expansion over the previous minima~\citep{mirzadeh2020understanding}.

We now proceed to discuss how several existing algorithms such as EWC~\citep{kirkpatrick2017overcoming}, Kronecker approximation~\citep{ritter2018online}, SI~\citep{zenke2017continual}, and MAS~\citep{aljundi2018memory} can be considered as variants of our framework. First, the constant term (a) in Eq.~\eqref{eq:loss_prox} does not depend of $w$ and thus can be ignored. Moreover, at the end of each task, the gradient term $\hatL_{k'}(\hatw_{k'})$ is usually small, so term (b) in Eq.~\eqref{eq:loss_prox} is usually skipped in practice. In the following, we show that several existing algorithms are basically using different approaches to approximate the Hessian matrix in term (c), and thus are special cases in our framework.

\paragraph*{Elastic Weight Consolidation (EWC)}
The EWC algorithm uses a diagonal matrix whose diagonal elements are the Fisher information of the model weights to construct the regularization term.
The intuition of EWC is that the weights with higher Fisher information are more important to current task, and thus should undergo less change in future tasks. As we know, if the loss function is the negative log-likelihood, and we obtained the ground truth probabilistic model, then the Fisher information matrix is equivalent to the Hessian matrix (see \eg Chapter 4.5 of~\cite{keener2011theoretical}). Hence, the EWC algorithm can be considered as a method that uses the diagonal elements of the Hessian matrix to approximate the full Hessian in Eq.~\eqref{eq:loss_prox}.

\paragraph*{Kronecker factored Laplace approximation}
The basic idea of the Kronecker factored Laplace approximation algorithm by~\cite{ritter2018online} is similar to that of EWC, with the key difference being that it uses Kronecker product approximation of the Fisher information matrix~\citep{martens2015optimizing,grosse2016kronecker}. In other words, this approximation takes the off-diagonal elements of the Hessian matrix into consideration, and thus usually performs better tha EWC.

Finally, the SI and MAS algorithms construct the diagonal regularization function based on the concept of intelligent synapses. Although the basic idea is different from Hessian/Fisher matrix approximation, these algorithms can also be analyzed in our framework as the regularization function is quadratic. More specifically, our analysis mainly relies on an assumption that the difference between the approximate Hessian matrix (the diagonal regularization matrix in SI and MAS) and the ground truth Hessian matrix is bounded in operator norm (see Assumption~\ref{asm:hess_prox} in Section~\ref{sec:analysis}), and our analysis can be applied to SI and MAS as long as this assumption is satisfied.

Equipped with the loss approximation viewpoint, we are ready to perform theoretical analysis of regularization-based algorithms in order to get deeper understanding of their behavior. We provide analyses from both optimization (Section~\ref{sec:analysis}) and generalization (Section~\ref{sec:generalization}) aspects.

\section{Optimization Analysis}\label{sec:analysis}

As we can see, the key idea in regularization-based algorithms is to approximate the loss functions of previous tasks using quadratic functions. This leads to the following theoretical question: \emph{By running gradient descent algorithm on an approximate loss function, can we still minimize the actual loss that we are interested in?} Intuitively, the answer to this question depends on several factors:
\begin{list}{\labelitemi}{\leftmargin=2em}
\item[(i)] Approximation error of the Hessian matrix. A more accurate approximation of the Hessian matrix leads to better approximation the loss function, and thus it becomes easier to minimize the actual loss given an approximation of it.
\item[(ii)] The moving distance in the parameter space. We perform Taylor expansion at the end of each task. This quadratic approximation can become more and more inaccurate as the model parameters move away from the end-of-task points. Therefore, if we only need to move a small distance between tasks, \ie tasks are similar, then we may be able to achieve better performance.
\item[(iii)] High order derivatives. According to the properties of Taylor expansion, the error between the Taylor approximation and the actual function depends on the high order derivatives. For quadratic approximation, when the third derivative of the loss function is small, it becomes easier to keep minimizing the actual loss.
\end{list}

In the following, we rigorously demonstrate these points. For the purpose of theoretical analysis, we make a few simplifications to our setup. Without loss of generality, we study the training process of the last task $\T_K$, and still use $\hatw_{k}$ to denote the model parameters that we obtain at the end of the $k$-th task. We use the loss function approximation in Eq.~\eqref{eq:loss_prox}, but in this section, for simplicity we ignore the finite-sample effect and replace the empirical loss function with the population loss function, \ie we define
\[
  \tildeL_{K-1}(w) = \sum_{k=1}^{K-1} L_{k}(\hatw_{k}) + (w - \hatw_{k})^\top\nabla L_{k}(\hatw_{k})
  + \frac{1}{2}(w - \hatw_{k})^\top H_{k} (w - \hatw_{k}),
\]
where $H_{k}$ represents $\nabla^2 L_{k}(\hatw_{k})$ or its approximation.\footnote{In other words, $\tildeL_{K-1}(w)$ is the \emph{population} version of $L_{K-1}^{\text{prox}}(w)$ in Section~\ref{sec:approach}.} We discuss the finite sample (generalization) effect in the next section. As mentioned, during the training of the last task, we have access to the approximate loss function
\[
\tildeF(w) := \frac{1}{K}(\tildeL_{K-1}(w) + L_K(w)),
\]
whereas the actual loss function that we care about is
\[F(w) := \frac{1}{K}\sum_{k=1}^{K} L_{k}(w).\]
We also focus on gradient descent instead of its stochastic counterpart. In particular, let $w_0:=\hatw_{K-1}$ be the initial model parameter for the last task. We run the following update for $t=1,2,\ldots, T$:
\begin{equation}\label{eq:gd_update}
    w_t = w_{t-1} - \eta \nabla \tildeF(w_{t-1}).
\end{equation}

We use the following standard notions for differentiable function $f:\W\mapsto\R$.
\begin{definition}\label{def:smoothness}
$f$ is $\mu$-smooth if $\twonms{\nabla f(w) - \nabla f(w')} \le \mu \twonms{w - w'}$, $\forall w,w'\in\W$.
\end{definition}
\begin{definition}\label{def:hess_lip}
$f$ is $\rho$-Hessian Lipschitz if $\twonms{\nabla^2 f(w) - \nabla^2 f(w')}\le \rho\twonms{w-w'}$, $\forall w,w'\in\W$.
\end{definition}

We make the assumptions that the loss functions are smooth and Hessian Lipschitz. We note that the Hessian Lipschitz assumption is standard in analysis of non-convex optimization~\citep{nesterov2006cubic,jin2017escape}, and that the Hessian Lipschitz constant is an upper bound for the third derivative.
\begin{asm}\label{asm:smooth_hess_lip}
We assume that $L_k(w)$ is $\mu$-smooth and $\rho$-Hessian Lipschitz $\forall k\in[K]$.
\end{asm}

We also assume that the error between the matrices $H_{k}$ and $\nabla^2 L_k(\hatw_k)$ is bounded.
\begin{asm}\label{asm:hess_prox}
We assume that for every $k\in[K]$, $\twonms{H_k} \le \mu$, where $\mu$ is defined in Assumption~\ref{asm:smooth_hess_lip}, and that $\twonms{H_k - \nabla^2 L_k(\hatw_k)}\le \delta$ for some $\delta \ge 0$.
\end{asm}

\subsection{Sufficient and Worst-Case Necessary Condition for One-Step Update}
We begin with analyzing a single step during training. Our goal is to understand by running a single step of gradient descent on $\tildeF(w)$, whether we can minimize the actual loss function $F(w)$. More specifically, we have the following result.
\begin{theorem}\label{thm:single_step}
Under Assumptions~\ref{asm:smooth_hess_lip} and~\ref{asm:hess_prox}, and suppose that in the $t$-th iteration, we observe
\begin{align}\label{eq:grad_norm_condi}
    \twonms{\nabla \tildeF(w_{t-1})} \ge  \frac{c}{K}\sum_{k=1}^{K-1}  \delta \twonms{w_{t-1} - \hatw_k}
    + \rho \twonms{w_{t-1} - \hatw_k}^2,
\end{align}
for some $c>1$, and the learning rate satisfies $\eta \le \frac{2(1-1/c)}{\mu}$, then we have
\[
F(w_t) \le F(w_{t-1}) - \eta(1 - \frac{1}{c} - \frac{\mu\eta}{2})\twonms{\nabla\tildeF(w_{t-1})}^2.
\]
\end{theorem}
We prove Theorem~\ref{thm:single_step} in Appendix~\ref{prf:single_step}. Here, we emphasize that this result does not assume any convexity of the loss functions. The theorem provides a sufficient condition Eq.~\eqref{eq:grad_norm_condi}, under which by running gradient descent on $\tildeF$, we can still minimize the true loss function $F$.
Intuitively, when the right hand side of Eq.~\eqref{eq:grad_norm_condi} is smaller, this condition is more likely to be satisfied. This implies that when (i) the approximation error of the Hessian matrix, \ie, $\delta$ is small, or (ii) the moving distance in the parameter space $\twonms{w_{t-1} - \hatw_k}$ is small or (iii) the third derivative $\rho$ is small, condition Eq.~\eqref{eq:grad_norm_condi} becomes easier to hold. Therefore, this theorem validates our hypotheses at the beginning of this section. In addition, we note that Theorem~\ref{thm:single_step} also implies that as training goes on and $\twonms{\nabla\tildeF(w_{t})}$ decreasing, it is beneficial to decrease the learning rate $\eta$, since when $c$ decreases, the upper bound on $\eta$ that guarantees the decay of $F$ (\ie $2(1-1/c)/\mu$) also decreases. We notice that the importance of learning rate decay for continual learning has been observed in some empirical study recently~\citep{mirzadeh2020dropout}.

In Proposition~\ref{prop:lower_bound} below, we will see that this condition is also necessary in the worst-case scenario, at least for the case where $K=2$. More specifically, we can construct cases in which~\eqref{eq:grad_norm_condi} is violated and the gradients of $F(w)$ and $\tildeF(w)$ have opposite directions.

\begin{prop}\label{prop:lower_bound}
Suppose that $K=2$, $d=1$, $\W = [0, 1]$. Then, there exists $\hatw_1$, $L_1(w)$, $\tildeL_1(w)$, and $L_2(w)$ such that if
$|\tildeF'(w)| < \frac{1}{2}[\delta|w - \hatw_1| + \rho(w - \hatw_1)^2]$, then $\tildeF'(w)\cdot F'(w) < 0$.
\end{prop}
We prove Proposition~\ref{prop:lower_bound} in Appendix~\ref{prf:lower_bound}.

\subsection{Convergence Analysis}\label{sec:convergence}

Although the condition in Eq.~\eqref{eq:grad_norm_condi} provides us with insights on the dynamics of the training algorithm, it is usually hard to check this condition in every step, since we may not have good estimates of $\delta$ and $\rho$. A practical implementation is to choose a constant learning rate along with an appropriate number of training steps. In this section, we provide bounds on the convergence behavior of our algorithm with a constant learning rate and $T$ iterations, both for non-convex and convex loss functions.
These results imply that sometimes early stopping may be helpful, especially when $\delta$ or $\rho$ is large or the tasks are more distinct from each other, \ie the moving distance between tasks is large.
We begin with a convergence analysis for non-convex loss functions in Theorem~\ref{thm:nonconvex}, in which we use the common choice of learning rate $1/\mu$ for gradient descent on smooth functions~\citep{bubeck2014convex}.

\begin{theorem}[non-convex]\label{thm:nonconvex}
Let $F_0:=F(w_0)$, $F^* := \min_{w\in\W}F(w)$, $\tildeF_0 := \tildeF(w_0)$, and $\tildeF^* := \min_{w\in\W}\tildeF(w)$. Then, under Assumptions~\ref{asm:smooth_hess_lip} and~\ref{asm:hess_prox}, after running $T$ iterations of the gradient descent update~\eqref{eq:gd_update} with learning rate $\eta = 1/\mu$, we have
\[
\frac{1}{T}\sum_{t=1}^T\twonms{\gradF(w_{t-1})} \le \frac{\alpha}{\sqrt{T}} + \beta + \gamma_1\sqrt{T} + \gamma_2T,
\]
where $\alpha = \sqrt{2\mu(F_0 - F^*)}$, $\beta = \frac{\sqrt{3}}{K}\sum_{k=1}^{K-2}\delta\twonms{w_0 - \hatw_k} + 2\rho\twonms{w_0 - \hatw_k}^2$, $\gamma_1 = \delta\sqrt{\frac{3}{\mu}(\tildeF_0 - \tildeF^*)}$, and $\gamma_2 = \frac{4\rho}{\mu}(\tildeF_0 - \tildeF^*)$.
\end{theorem}
We prove Theorem~\ref{thm:nonconvex} in Appendix~\ref{prf:nonconvex}. Unlike standard optimization analysis, the average norm of the gradients does not always decrease as $T$ increases, when $\delta\neq 0$ or $\rho \neq 0$. As we can see, the coefficients $\gamma_1$ and $\gamma_2$ in the last two terms that ``hurt'' convergence increase as $\delta$ and $\rho$ increase, respectively.


When the loss functions are convex, we can prove a better guarantee which does not have the $\bigo(\sqrt{T})$ and $\bigo(T)$ terms as in Theorem~\ref{thm:nonconvex}. More specifically, we have the following assumption and theorem.
\begin{asm}\label{asm:convexity}
$L_k(w)$ is convex and $H_k \succeq 0$, $\forall k$. 
\end{asm}

\begin{theorem}[convex]\label{thm:convex}
Suppose that Assumptions~\ref{asm:smooth_hess_lip},~\ref{asm:hess_prox}, and~\ref{asm:convexity} hold, and define $F^* = \min_{w\in\W}F(w)$, $w^*\in\arg\min_{w\in\W}F(w)$, $\widetilde{w}^*\in\arg\min_{w\in\W}\tildeF(w)$, and $\tildeD:=\twonms{w_0 - \widetilde{w}^*}$. After running $T$ iterations of the gradient descent update~\eqref{eq:gd_update} with learning rate $\eta = 1/\mu$, we have
\[
F(w_T) - F^* \le \frac{\alpha}{T} + \beta,
\]
where $\alpha = 2\mu \tildeD^2$, and
$\beta = \frac{1}{2K}\sum_{k=1}^{K-1} \delta(\twonms{{w}^* - \hatw_k}^2 + 2\tildeD^2 + 2\twonms{\widetilde{w}^* - \hatw_k}^2) + \rho(\twonms{{w}^* - \hatw_k}^3 + 4\tildeD^3 + 4\twonms{\widetilde{w}^* - \hatw_k}^3)$.
\end{theorem}
We prove Theorem~\ref{thm:convex} in Appendix~\ref{prf:convex}. As we can see, if $\delta\neq 0 $ or $\rho \neq 0$, we still cannot guarantee the convergence to the true minimum of $F$, due to the inexactness of $\tildeF$. On the other hand, if the loss functions are quadratic and we save the full Hessian matrices, \ie $\delta=\rho=0$, as we have full information about previous loss functions, we can recover the standard $\bigo(1/T)$ convergence rate for gradient descent on convex and smooth functions.

\section{Generalization Analysis}\label{sec:generalization}

While the previous section provides with important insights on the convergence behavior of the algorithm, one aspect that is less studied is the finite-sample analysis of continual learning. More specifically, what we eventually care about is the average \emph{population} loss across all the tasks, \ie $F(w) = \frac{1}{K}\sum_{k=1}^{K} L_{k}(w)$, but what we have at the end of the last task is the \emph{empirical} loss of the last task, \ie $\widehat{L}_K(w) := \frac{1}{n_K}\sum_{i=1}^{n_K} \ell_K(w; x_{K,i}, y_{K,i})$, as well as the \emph{empirical quadratic approximation}:
\begin{equation}\label{eq:approx_empirical}
L^\prox_{K-1}(w) = \sum_{k=1}^{K-1} \hatL_{k}(\hatw_{k}) + (w - \hatw_{k})^\top\nabla \hatL_{k}(\hatw_{k}) \nonumber + \frac{1}{2}(w - \hatw_{k})^\top H_{k} (w - \hatw_{k}).
\end{equation}
Thus, our generalization analysis aims to bound the gap between $F(w)$ and $\frac{1}{K}(\widehat{L}_K(w) + L^\prox_{K-1}(w))$. Since our focus in this section is the finite-sample effect, we will not consider the approximation error of the empirical Hessian matrix, and instead, we assume that $H_k$ here is the exact empirical Hessian matrix. The extension of our results to an approximate empirical Hessian matrix is straightforward.
\begin{asm}\label{asm:empirical_hess}
In Eq.~\eqref{eq:approx_empirical}, $H_{k}=\nabla^2\hatL_k(\hatw_k)=\frac{1}{n_k}\sum_{i=1}^{n_k}\nabla^2\ell(\hatw_k; x_{k,i}, y_{k,i})$.
\end{asm}
In addition, we make the standard boundedness assumptions that are commonly used in uniform convergence analysis~\citep{mohri2018foundations,foster2018uniform}.
\begin{asm}\label{asm:boundedness}
For any $k\in[K]$, $w\in\W$ and $(x, y) \in \X \times \Y$, $|\ell_k(w;x, y)| \le b$, $\twonms{\nabla \ell_k(w;x, y)} \le g$, and $\twonms{\nabla^2 \ell_k(w;x,y)} \le h$.
\end{asm}
With these assumptions, we are ready to state our generalization bound.
\begin{theorem}\label{thm:generalization}
Under Assumptions~\ref{asm:empirical_hess} and~\ref{asm:boundedness}, with probability at least $1-\delta$ over the random training examples $(x_{k,i}, y_{k,i})$, $i\in[n_k]$, $k\in[K]$, we have
\begin{align*}
    F(w) \le & \underbrace{\frac{1}{K}(\widehat{L}_K(w) + L^{\text{\rm prox}}_{K-1}(w))}_{\text{\rm regularized training loss}} + \underbrace{\frac{2}{K}\Big(bR^{(\ell)}_K  +  \sum_{k=1}^{K-1}(bR_k^{(\ell)} + g R_k^{(g)} +hR_k^{(h)})\Big) + \bigo(\frac{\log(\frac{K}{\delta})}{\sqrt{n}})}_{\text{\rm finite-sample effect}} +\underbrace{\frac{\rho}{2K}\sum_{k=1}^{K-1}\twonms{w - \hatw_k}^3}_{\text{\rm loss approximation error}},
\end{align*}
where
\begin{align*}
    R^{(\ell)}_k &= \EE_\sigma\Big[\frac{1}{n_k}\sup_{w}\sum_{i=1}^{n_k}\sigma_i \ell_k(w; x_{k,i}, y_{k,i})\Big], \\
    R^{(g)}_k &= \EE_{\sigma}\Big[ \frac{1}{n_k}\sup_{w}\twonms{\sum_{i=1}^{n_k} \sigma_i \nabla\ell_k(w;x_{k,i}, y_{k,i})}\Big], \\ R^{(h)}_k &=\EE_{\sigma}\Big[\frac{1}{n_k} \sup_{w}\twonms{\sum_{i=1}^{n_k} \sigma_i \nabla^2 \ell_k(w;x_{k,i}, y_{k,i})}\Big]
\end{align*} are the (normed) Rademacher complexities for the value, gradient and Hessian matrix of the loss functions ($\sigma$ denotes a sequence of i.i.d. Rademacher random variables); and $n=\min_{k\in[K]}n_k$.
\end{theorem}
We prove Theorem~\ref{thm:generalization} in Appendix~\ref{prf:generalization}. As we can see, the gap between $F(w)$ and the regularized training loss consists of two parts. The first part is related to the Rademacher complexity terms, which represent the finite-sample effect, and can approach zero as the number of training data for each task increases. The second part is the loss function approximation error, which does not decay when the number of training examples increases. This again demonstrates that the accuracy of the loss function approximation plays a key role in regularization-based continual learning.

\section{Experiments}\label{sec:experiments}
In this section, we present experimental results validating our theoretical findings.
More specifically, our experiments mainly validate two claims: (i) an accurate approximation of the Hessian matrix leads to better overall performance and (ii) training a large number of steps may not always be helpful, and sometimes early stopping is needed.
All the experiments are implemented with JAX~\citep{jax2018github}.

\subsection{Experiment Setup}

\paragraph*{Datasets.}
We use three standard continual learning benchmarks created based on MNIST~\citep{lecun1998gradient} and CIFAR-100~\citep{krizhevsky2009learning} datasets, \ie Permuted MNIST~\citep{goodfellow2013empirical}, Rotated MNIST~\citep{lopez2017gradient}, and Split CIFAR~\citep{chaudhry2018efficient}. In Permuted MNIST, for each task, we choose a random permutation of the pixels of MNIST images, and reorder all the images according to the permutation. We use $20$-task Permuted MNIST in the experiments. In Rotated MNIST, for each task, we rotate the MNIST images by a particular angle. In our experiments, we choose a $20$-task Rotated MNIST, with the rotation angles being $0,10,20,\ldots,190$ degrees. For Split CIFAR, we split the $100$ labels of the CIFAR-100 dataset to $20$ disjoint subsets, each subset corresponding to one coarse label.\footnote{CIFAR-100 contains $20$ coarse labels, and each of them contains $5$ fine-grained labels.} For each task, we use the CIFAR-100 data whose labels belong to a particular subset.

\paragraph*{Architecture.}
For Permuted MNIST and Rotated MNIST, we use a multilayer perceptron (MLP) with two hidden layers, each having $256$ units. For Split CIFAR, we use a convolutional neural network (CNN) with two convolutional layers with $3\times 3$ kernel and $64$ and $128$ output channels, respectively. We add max pooling with stride $2$ after each convolutional layer. We add one hidden fully connected layer with $512$ units. Moreover, for this model, we use a multi-head structure similar to what has been used by~\cite{chaudhry2018efficient,farajtabar2019orthogonal}: instead of having $100$ logits in the output layer, we use separate heads for different tasks, and each head corresponds to the classes of the associated task. During training, for each task, we only optimize the cross-entropy loss over the logits and labels of the corresponding output head.
For both MLP and CNN models, we use ReLU activation function.

\paragraph*{Algorithms.}
We implement three regularization-based algorithms, namely EWC, Synaptic Intelligence (SI), and Kronecker factored Laplace approximation (Kronecker for short), which can be analyzed in our framework. We omit the constant and gradient terms ((a) and (b) in Eq.~\eqref{eq:loss_prox}), and following~\cite{kirkpatrick2017overcoming,zenke2017continual,ritter2018online}, we add an extra coefficient $\lambda$ in the quadratic regularization function as a hyper parameter, \ie we minimize $\frac{1}{k+1}(\hatL_{k+1}(w) + \lambda L_k^{\text{prox}}(w))$. In our experiments, we tune $\lambda$ and report the results with the choice of $\lambda$ that produces the best average test accuracy over all tasks. The hyper parameter choice for all the results are provided in Appendix~\ref{sec:experiment_detail}.  Following prior works~\citep{kirkpatrick2017overcoming,chaudhry2018efficient,farajtabar2019orthogonal}, we choose a learning rate of $10^{-3}$ and a batch size of $10$. For all the results that we report, we present the average result over $5$ independent runs, as well as the standard deviation.

\subsection{Results}

We evaluate the EWC, SI and Kronecker algorithms on the three benchmarks. As mentioned, the Kronecker approximation algorithm makes use of the off-diagonal elements of the Hessian matrix, and thus provides a better approximation compared to EWC and SI which use diagonal quadratic regularization. We note that~\cite{ritter2018online} have already observed that the Kronecker algorithm outperforms EWC and SI, and presented results on Permuted MNIST. Here, we include Permuted MNIST for completeness and also provide additional results on Rotated MNIST and Split CIFAR. The average test accuracy over the $20$ tasks are presented in Tables~\ref{tab:permuted_mnist},~\ref{tab:rotated_mnist}, and~\ref{tab:split_cifar} for the three benchmarks, respectively. In these tables, we also provide the results for the \emph{vanilla SGD} algorithm, which sequentially trains over all tasks without storing any side information. In the caption of each table, we also provide the average test accuracy for the \emph{multi-task} algorithm, which has access to all the data from all the tasks simultaneously. In all of these tables, we present the results as a function of the number of epochs that we train for each task. As we can see, Kronecker algorithm consistently performs the best. These results confirm our theoretical finding that an accurate Hessian approximation is important for regularization-based algorithms.

\begin{table}[ht]
\caption{Average test accuracy on Permuted MNIST. Test accuracy for multi-task algorithm is $96.8 \pm 0.0$.}\label{tab:permuted_mnist}
\centering
\setlength\tabcolsep{9pt}
\begin{tabular}{c|cccc}
\toprule
	 & \multicolumn{4}{c}{\bf Average accuracy $\pm$ Std. ($\%$)} \\
epoch  &  Vanilla &  EWC & SI & Kronecker	\\
\midrule
	1 & $59.1 \pm 0.9$	& $62.8 \pm 0.7$ & $62.2 \pm 0.9$ &	$\mathbf{82.9 \pm 0.2}$ \\
	2 & $56.8 \pm 0.9$	& $59.8 \pm 0.4$ & $63.7 \pm 0.7$ &	$\mathbf{86.6 \pm 0.2}$\\
	4 & $57.1 \pm 0.6$	& $57.4 \pm 0.7$ & $69.4 \pm 0.3$	& $\mathbf{90.1 \pm 0.1}$ \\
	8 & $55.5 \pm 1.1$	& $56.1 \pm 0.7$ & $76.5 \pm 0.5$	& $\mathbf{93.0 \pm 0.1}$\\
	16 & $54.8 \pm 1.4$	& $56.3 \pm 1.3$ & $80.7 \pm 0.5$	& $\mathbf{94.8 \pm 0.1}$\\
	32 & $52.6 \pm 1.6$	& $56.9 \pm 1.9$ & $78.2 \pm 1.2$	& $\mathbf{96.0 \pm 0.0}$\\
\bottomrule
\end{tabular}
\end{table}

\begin{table}[ht]
\caption{Average test accuracy on Rotated MNIST. Test accuracy for multi-task algorithm is $97.2 \pm 0.1$.}\label{tab:rotated_mnist}
\centering
\setlength\tabcolsep{9pt}
\begin{tabular}{c|cccc}
\toprule
	 & \multicolumn{4}{c}{\bf Average accuracy $\pm$ Std. ($\%$)} \\
epoch  &  Vanilla &  EWC & SI & Kronecker	\\
\midrule
	1 & $41.9 \pm 0.6$	& $47.7 \pm 0.5$ & $42.1 \pm 0.3$ & $\mathbf{69.1 \pm 0.5}$ \\
	2 & $43.5 \pm 0.6$	& $50.2 \pm 0.4$ & $44.3 \pm 0.3$	& $\mathbf{70.0 \pm 0.3}$\\
	4 & $44.6 \pm 0.3$	& $53.9 \pm 0.9$ &  $47.4 \pm 0.3$ &	$\mathbf{71.9 \pm 0.7}$ \\
	8 & $45.7 \pm 0.5$	& $57.9 \pm 1.4$ & $50.5 \pm 0.3$  &	$\mathbf{75.5 \pm 0.8}$\\
	16 & $46.2 \pm 0.5$	& $60.9 \pm 1.4$ & $54.8 \pm 0.6$ &	$\mathbf{78.8 \pm 0.8}$\\
	32 & $46.6 \pm 0.5$	& $62.4 \pm 2.5$ & $58.3 \pm 0.8$  &	$\mathbf{81.9 \pm 0.9}$\\
\bottomrule
\end{tabular}
\end{table}

\begin{table}[ht]
\caption{Average test accuracy on Split CIFAR. Test accuracy for multi-task algorithm is $62.6 \pm 0.2$.}\label{tab:split_cifar}
\centering
\setlength\tabcolsep{9pt}
\begin{tabular}{c|cccc}
\toprule
	 & \multicolumn{4}{c}{\bf Average accuracy $\pm$ Std. ($\%$)} \\
epoch  &  Vanilla &  EWC &  SI & Kronecker	\\
\midrule
	1 & $29.6 \pm 0.4$ &	$29.4 \pm 1.3$ & $29.9 \pm 0.9$ &	$\mathbf{31.6 \pm 0.8}$ \\
	2 & $32.6 \pm 1.1$	& $32.6 \pm 0.9$ & $32.8 \pm 0.7$	& $ \mathbf{37.2 \pm 0.6}$\\
	4 & $35.3 \pm 0.7$	& $34.9 \pm 0.8$ & $35.1 \pm 1.0$	& $\mathbf{42.0 \pm 0.7}$ \\
	8 & $39.4 \pm 1.6$	& $38.4 \pm 1.0$ & $38.7 \pm 1.3$	& $\mathbf{47.3 \pm 0.8}$\\
	16 & $43.2 \pm 1.4$	& $41.3 \pm 1.6$ & $43.2 \pm 0.5$	& $\mathbf{54.2 \pm 0.7}$\\
	32 & {$\it{38.0 \pm 0.6}$}	& $44.0 \pm 0.9$ &  $45.3 \pm 0.7$ & $\mathbf{59.9 \pm 0.7}$\\
	64 & $\it{32.4 \pm 0.2}$ & $\it{40.9 \pm 0.7}$ & $\it{45.1 \pm 1.1}$ & \textit{\textbf{59.7}} $\pm$ \textit{\textbf{0.6}}\\
    128 & $\it{30.9 \pm 1.6}$ &	$\it{37.0 \pm 0.9}$ & $\it{43.8 \pm 0.8}$ & \textit{\textbf{53.1}} $\pm$ \textit{\textbf{1.1}} \\
\bottomrule
\end{tabular}
\end{table}

Another interesting observation is that on Permuted MNIST, for the vanilla SGD and EWC algorithms, the average test accuracy \emph{decreases} as we \emph{increase} the number of epochs to train for each task. Similar phenomenon also happens for vanilla SGD, EWC and SI algorithms on Split CIFAR, when we increase the number of epochs per task beyond $32$ (see the numbers in \emph{italic} font in Table~\ref{tab:split_cifar}). This validates our theoretical results that training longer on each individual task may not necessarily be helpful for achieving a good final average accuracy, especially when the approximation error for the Hessian matrix is large. In fact, the behavior of vanilla SGD, EWC, and SI on Split CIFAR exactly matches the prediction from Theorem~\ref{thm:nonconvex}: When we increase the number of epochs per task, the average test accuracy first improves (since the model is trained better for each task), and then becomes worse as the model forgets early tasks.

As a final remark, we note that the strong performance of the Kronecker algorithm comes with a large computation cost. We observe that in our experiments on Permuted/Rotated MNIST, the computation time of Kronecker algorithm is $5$ times as much as that of the EWC algorithm, and on Split CIFAR, this ratio is $10$. Therefore, for practical applications, we may need to take the trade-off between the accuracy of Hessian approximation and the computational cost into consideration.

\section{Conclusions}
We studied a unified framework for regularization-based continual learning through the lens of loss function approximation. We showed that existing algorithms such as EWC and Kronecker approximation are special cases of our framework. We further established theoretical guarantees for the algorithms on both optimization and generalization aspects.
To the best of our knowledge, our work is among the first ones that try to formulate continual learning from a loss function approximation perspective that enables studying its convergence behavior and finite-sample guarantees.
Finally, we validated the theoretical findings with experiments on several continual learning benchmarks.

\section*{Acknowledgements}

We would like to thank Dilan Gorur, Clara Huiyi Hu, Nevena Lazic, and Michalis Titsias for helpful discussions.

\bibliographystyle{abbrvnat}
\bibliography{reference.bib}
\newpage
\appendix
\section*{Appendix}
\section{Proof of Theorem~\ref{thm:single_step}}\label{prf:single_step}
We first provide a bound for the difference between the gradients of $\nabla \tildeF(w)$ and $\nabla F(w)$.
\begin{lemma}\label{lem:bound_delta}
Let $\Delta(w) = \nabla \tildeF(w) - \nabla F(w)$. Then we have
\[
\twonms{\Delta(w)} \le \frac{1}{K}\sum_{k=1}^{K-1}\delta \twonms{w - \hatw_k} + \rho\twonms{w - \hatw_k}^2.
\]
\end{lemma}
We prove Lemma~\ref{lem:bound_delta} in Appendix~\ref{prf:bound_delta}. Since the loss functions for all the tasks $L_k(w)$ are $\mu$-smooth, we know that $F(w)$ is also $\mu$-smooth. Then we have
\begin{align*}
    F(w_t) &\le F(w_{t-1}) + \innerps{\nabla F(w_{t-1})}{w_t - w_{t-1}} + \frac{\mu}{2}\twonms{w_t - w_{t-1}}^2  \\
    &= F(w_{t-1}) + \innerps{\nabla \tildeF(w_{t-1}) - \Delta(w_{t-1})}{-\eta \nabla\tildeF(w_{t-1})} + \frac{\mu\eta^2}{2}\twonms{\nabla\tildeF(w_{t-1})}^2  \\
    &\le F(w_{t-1}) - \eta (1-\frac{\mu\eta}{2}) \twonms{\nabla \tildeF(w_{t-1})}^2 + \eta \twonms{\nabla\tildeF(w_{t-1})}\twonms{\Delta(w_{t-1})}.
\end{align*}
Therefore, as long as $\twonms{\nabla\tildeF(w_{t-1})} \ge c \twonms{\Delta(w_{t-1})}$ for some $c > 1$, we have
\begin{align}\label{eq:func_val_decay}
    F(w_t) \le F(w_{t-1}) - \eta(1 - \frac{1}{c} - \frac{\mu\eta}{2})\twonms{\nabla\tildeF(w_{t-1})}^2.
\end{align}
Then we can complete the proof by combining~\eqref{eq:func_val_decay} with Lemma~\ref{lem:bound_delta}.

\subsection{Proof of Lemma~\ref{lem:bound_delta}}\label{prf:bound_delta}
By the definition of $\tildeF(w)$, for some $\xi_k\in[0, 1]$, $k\in[K-1]$, we have
\begin{align*}
    \Delta(w) &= \frac{1}{K}\sum_{k=1}^{K-1} \nabla L_{k}(\hatw_{k}) + H_{k}(w - \hatw_{k}) - \nabla L_{k}(w)  \\
    &=\frac{1}{K}\sum_{k=1}^{K-1}\left(H_{k} - \nabla^2 L_k(\hatw_{k} + \xi_k(w - \hatw_k))\right)(w-\hatw_k),
\end{align*}
where the second equality is due to Lagrange's mean value theorem. Then, according to Assumptions~\ref{asm:smooth_hess_lip} and~\ref{asm:hess_prox}, we have
\begin{align}
   \twonm{H_{k} - \nabla^2 L_k(\hatw_{k} + \xi_k(w - \hatw_k))} &\le \delta + \twonm{\nabla^2 L_k(\hatw_k) - \nabla^2 L_k(\hatw_{k} + \xi_k(w - \hatw_k))}  \nonumber \\
   &\le \delta + \rho\twonms{w - \hatw_k}. \label{eq:hess_diff}
\end{align}
Then, according to triangle inequality, we obtain
\[
\twonms{\Delta(w)} \le \frac{1}{K}\sum_{k=1}^{K-1}\delta \twonms{w - \hatw_k} + \rho\twonms{w - \hatw_k}^2.
\]

\section{Proof of Proposition~\ref{prop:lower_bound}}\label{prf:lower_bound}
We first note that it suffices to construct $F(w)$ and $\tildeF(w)$, as one can always choose $L_2(w)\equiv 0$ and then the construction of $F(w)$ and $\tildeF(w)$ is equivalent to that of $L_1(w)$ and $\tildeL_1(w)$.
Let $\hatw_1 = 0$,
\begin{align*}
    F(w) &= (w-1)^2  + \frac{\rho}{6}w^3, w\in[0, 1], \\
    \tildeF(w) &= (w-1)^2 - \frac{\delta}{4}w^2, w\in[0, 1].
\end{align*}
One can easily check that $F(\hatw_1) = \tildeF(\hatw_1)$, $F'(\hatw_1) = \tildeF'(\hatw_1)$, and $|F''(\hatw_1)-\tildeF''(\hatw_1) | = \frac{\delta}{2}$. In addition, since the second derivative of $F(w)$ is always bounded in $[0, 1]$, we know that $F(w)$ is smooth. Since $F'''(w) \equiv \rho$, we know that $F(w)$ is $\rho$-Hessian Lipschitz. Therefore, $F(w)$ and $\tildeF(w)$ satisfy all of our assumptions.

Since $\tildeF'(w) = 2(w-1) -\frac{\delta}{2}w$, we know that $\tildeF'(w) < 0$, $\forall w\in[0, 1]$, and then
\[
|\tildeF'(w) | <\frac{\delta}{2}w + \frac{\rho}{2}w^2
\]
is equivalent to $\frac{\delta}{2}w - 2(w-1) < \frac{\delta}{2}w + \frac{\rho}{2}w^2$, which implies that $F(w) = 2(w-1) + \frac{\rho}{2}w^2 > 0$.

\section{Proof of Theorem~\ref{thm:nonconvex}}\label{prf:nonconvex}

Similar to Appendix~\ref{prf:single_step}, we define $\Delta(w)=\nabla \tildeF(w) - \nabla F(w)$. According to Assumptions~\ref{asm:smooth_hess_lip} and~\ref{asm:hess_prox}, we know that both $F(w)$ and $\tildeF$ are $\mu$-smooth. By the smoothness of $F(w)$ and using the fact that $\eta = 1/\mu$, we get
\begin{align}
    F(w_t) &\le F(w_{t-1}) + \innerps{\gradF (w_{t-1})}{w_{t} - w_{t-1}} + \frac{\mu}{2}\twonm{w_t - w_{t-1}}^2  \nonumber \\
    &=F(w_{t-1}) - \innerps{\gradF(w_{t-1})}{\eta(\gradF(w_{t-1}) + \Delta(w_{t-1}))} + \frac{\mu\eta^2}{2}\twonms{\gradF(w_{t-1}) + \Delta(w_{t-1})}^2  \nonumber \\
    &= F(w_{t-1}) - \frac{1}{2\mu} \twonms{\gradF(w_{t-1})}^2 + \frac{1}{2\mu}\twonms{\Delta(w_{t-1})}^2, \nonumber
\end{align}
which implies
\begin{align}\label{eq:fsmooth}
\twonms{\gradF(w_{t-1})}^2 \le 2\mu(F(w_{t-1}) - F(w_t)) + \twonms{\Delta(w_{t-1})}^2.
\end{align}
By averaging~\eqref{eq:fsmooth} over $t=1,\ldots, T$, we get
\[
\frac{1}{T}\sum_{t=1}^T \twonms{\gradF(w_{t-1})}^2 \le \frac{2\mu(F_0 - F^*)}{T} + \frac{1}{T}\sum_{t=1}^T \twonms{\Delta(w_{t-1})}^2.
\]
By taking square root on both sizes, and using Cauchy-Schwarz inequality as well as the fact that $\sqrt{a+b} \le \sqrt{a} + \sqrt{b}$, we get
\begin{align}\label{eq:fbound_1}
    \frac{1}{T}\sum_{t=1}^T\twonms{\gradF(w_{t-1})} \le \frac{\sqrt{2\mu(F_0 - F^*)}}{\sqrt{T}} + \sqrt{\frac{1}{T}\sum_{t=1}^T\twonms{\Delta(w_{t-1})}^2}.
\end{align}
We then proceed to bound $\twonms{\Delta(w_{t-1})}^2$. According to Lemma~\ref{lem:bound_delta}, we have
\begin{align*}
    \twonms{\Delta(w_{t-1})} &\le \frac{1}{K}\sum_{k=1}^{K-1}\delta(\twonms{w_{t-1} - w_0} + \twonms{w_0 - \hatw_k}) + 2\rho(\twonms{w_{t-1} - w_0}^2 + \twonms{w_0 - \hatw_k}^2) \nonumber \\
    &:= C + \delta\twonms{w_{t-1} - w_0} + 2\rho\twonms{w_{t-1} - w_0}^2,
\end{align*}
where $C:=\frac{1}{K}\sum_{k=1}^{K-2}\delta\twonms{w_0 - \hatw_k} + 2\rho\twonms{w_0 - \hatw_k}^2$ does not depend on the iteration count $t$. By Cauchy-Schwarz inequality, we get
\begin{align}\label{eq:delta_bound}
    \twonms{\Delta(w_{t-1})}^2 \le 3C^2 + 3\delta^2\twonms{w_{t-1} - w_0}^2 + 12\rho^2\twonms{w_{t-1} - w_0}^4.
\end{align}
Then we bound $\twonms{w_{t-1} - w_0}^2$. By trianlge inequality and Cauchy-Schwarz inequality, we obtain
\[
\twonms{w_{t-1} - w_0} = \twonm{-\eta\sum_{\tau=0}^{t-2} \nabla\tildeF(w_\tau)}
\le \eta \sum_{\tau=0}^{t-2}\twonms{\nabla\tildeF(w_\tau)}
\le \eta \sqrt{(t-1)\sum_{\tau=0}^{t-2} \twonms{\nabla\tildeF(w_{\tau})}^2 },
\]
and therefore
\begin{align}\label{eq:bound_w_change}
    \twonms{w_{t-1} - w_0}^2 \le \eta^2(t-1)\sum_{\tau=0}^{t-2}\twonms{\nabla\tildeF(w_\tau)}^2.
\end{align}
On the other hand, since $\tildeF(w)$ is also $\mu$-smooth, we have for every $t\ge 1$,
\begin{align*}
    \tildeF(w_t) &\le \tildeF(w_{t-1}) + \innerps{\nabla\tildeF(w_{t-1})}{w_t - w_{t-1}} + \frac{\mu}{2}\twonms{w_t - w_{t-1}}^2 \\
    &= \tildeF(w_{t-1}) - \frac{1}{2\mu}\twonms{\nabla\tildeF(w_{t-1})}^2,
\end{align*}
where in the equality we use the fact that $\eta = 1/\mu$. This implies that
\begin{align}\label{eq:grad_sum}
    \sum_{\tau=0}^{t-2} \twonms{\nabla\tildeF(w_{t-1})}^2 \le 2\mu(\tildeF_0 - \tildeF^*).
\end{align}
By combining~\eqref{eq:bound_w_change} and~\eqref{eq:grad_sum}, we obtain
\begin{align}\label{eq:bound_w_change_2}
    \twonms{w_{t-1} - w_0}^2 \le \frac{2}{\mu}(\tildeF_0 - \tildeF^*)(t-1),
\end{align}
and combining~\eqref{eq:delta_bound} and~\eqref{eq:bound_w_change_2}, we get
\begin{align}\label{eq:delta_bound_2}
    \twonms{\Delta(w_{t-1})}^2 \le 3C^2 + \frac{6\delta^2}{\mu}(\tildeF_0 - \tildeF^*)(t-1) + \frac{48\rho^2}{\mu^2}(\tildeF_0 - \tildeF^*)^2(t-1)^2.
\end{align}
By averaging~\eqref{eq:delta_bound_2} over $t=1,\ldots,T$ and plugging the result in~\eqref{eq:fbound_1}, we obtain
\[
\frac{1}{T}\sum_{t=1}^T\twonms{\gradF(w_{t-1})} \le \frac{\sqrt{2\mu(F_0 - F^*)}}{\sqrt{T}} + \sqrt{3}C + \delta\sqrt{\frac{3}{\mu}(\tildeF_0 - \tildeF^*)}\sqrt{T} + \frac{4\rho}{\mu}(\tildeF_0 - \tildeF^*)T,
\]
which completes the proof.

\section{Proof of Theorem~\ref{thm:convex}}\label{prf:convex}
Let $\tildeF^* = \min_{w\in\W}\tildeF(w)$. Since we run gradient descent with learning rate $\eta = 1/\mu$ on the convex and $\mu$-smooth function $\tildeF$, according to standard results in convex optimization~\cite{bubeck2014convex}, we have
\begin{align}\label{eq:cvx_tildef}
    \tildeF(w_T) - \tildeF^* \le \frac{2\mu \tildeD^2}{T}.
\end{align}
We provide the following lemma that bounds the difference between $\tildeF(w)$ and $F(w)$.
\begin{lemma}\label{lem:diff_function_val}
For any $w\in\W$, we have
\[
|F(w) - \tildeF(w)| \le \frac{1}{2K}\sum_{k=1}^{K-1}\delta \twonms{w - \hatw_k}^2 + \rho\twonms{w - \hatw_k}^3
\]
\end{lemma}
We prove Lemma~\ref{lem:diff_function_val} in Appendix~\ref{prf:diff_function_val}. Here, we proceed to analyze $F(w_T) - F^*$. We have
\begin{align}
    F(w_T) - F^* &= F(w_T) - \tildeF(w_T) + \tildeF(w_T) - \tildeF^* + \tildeF^* - F^* \nonumber \\
    &\le F(w_T) - \tildeF(w_T) + \tildeF^* - F^* + \frac{2\mu \tildeD^2}{T}. \label{eq:split_loss}
\end{align}
To bound $F(w_T) - \tildeF(w_T)$, we use the fact that for any convex and smooth functions, when we run gradient descent with learning rate $1/\mu$, the iterates only move closer to the minimum of the function, \ie $\twonms{w_T - \widetilde{w}^*} \le \tildeD$. Then, by triangle inequality, for any $k\in[K-1]$
\[
\twonms{w_T - \hatw_k} \le \twonms{w_T - \widetilde{w}^*} + \twonms{\widetilde{w}^* - \hatw_k} \le \tildeD + \twonms{w_0 - \hatw_k}.
\]
Using the fact that for any two positive numbers $a$ and $b$, $(a+b)^2 \le 2(a^2+b^2)$ and $(a+b)^3 \le 4(a^3 + b^3)$, we obtain
\begin{align}
    \twonms{w_T - \hatw_k}^2 &\le 2(\tildeD^2 + \twonms{\widetilde{w}^* - \hatw_k}^2),  \label{eq:bound_wt_2} \\
    \twonms{w_T - \hatw_k}^3 &\le 4(\tildeD^3 + \twonms{\widetilde{w}^* - \hatw_k}^3).  \label{eq:bound_wt_3}
\end{align}
By combining~\eqref{eq:bound_wt_2} and~\eqref{eq:bound_wt_3} with Lemma~\ref{lem:diff_function_val}, we obtain
\begin{align}\label{eq:bound_fwt}
    F(w_T) - \tildeF(w_T) \le \frac{1}{K}\sum_{k=1}^{K-1} \delta(\tildeD^2 + \twonms{\widetilde{w}^* - \hatw_k}^2) + 2\rho(\tildeD^3 + \twonms{\widetilde{w}^* - \hatw_k}^3).
\end{align}
We can use a similar argument to bound $\tildeF^* - F^*$. Note that $\tildeF^* - F^* \le \tildeF(w^*) - F(w^*)$. Therefore, according to Lemma~\ref{lem:diff_function_val}, we have
\begin{align}\label{eq:bound_fstar}
    \tildeF^* - F^* \le \frac{1}{2K}\sum_{k=1}^{K-1}\delta \twonms{w^* - \hatw_k}^2 + \rho\twonms{w^* - \hatw_k}^3.
\end{align}
Then we can complete the proof by combining~\eqref{eq:split_loss},~\eqref{eq:bound_fwt}, and~\eqref{eq:bound_fstar}.

\subsection{Proof of Lemma~\ref{lem:diff_function_val}}\label{prf:diff_function_val}
By definition of $F(w)$ and $\tildeF(w)$, we have
\begin{align*}
    F(w) - \tildeF(w) =& \frac{1}{K}\sum_{k=1}^{K-1} L_k(w) - \left( L_k(\hatw_k) + (w - \hatw_k)^\top \nabla L_k(\hatw_k) + \frac{1}{2}(w - \hatw_k)^\top H_k (w - \hatw_k) \right) \\
    =&\frac{1}{K}\sum_{k=1}^{K-1}\frac{1}{2}(w-\hatw_k)^\top(\nabla^2L_k(\hatw_{k} + \xi_k(w - \hatw_k)) - H_k)(w-\hatw_k),
\end{align*}
for some $\xi_k\in[0,1]$. Then according to~\eqref{eq:hess_diff}, we get
\begin{align*}
    |F(w) - \tildeF(w)| &\le \frac{1}{2K}\sum_{k=1}^{K-1}\twonms{\nabla^2L_k(\hatw_{k} + \xi_k(w - \hatw_k)) - H_k} \twonms{w - \hatw_k}^2  \\
    &\le \frac{1}{2K}\sum_{k=1}^{K-1}\delta \twonms{w - \hatw_k}^2 + \rho\twonms{w - \hatw_k}^3.
\end{align*}

\section{Proof of Theorem~\ref{thm:generalization}}\label{prf:generalization}
First, recall the definitions of the population quadratic approximation function $\tildeL_{K-1}(w)$ and the empirical quadratic approximation function $L^\prox_{K-1}(w)$:
\begin{align*}
\tildeL_{K-1}(w) &= \sum_{k=1}^{K-1} L_{k}(\hatw_{k}) + (w - \hatw_{k})^\top\nabla L_{k}(\hatw_{k})
+ \frac{1}{2}(w - \hatw_{k})^\top \nabla^2 L_{k}(\hatw_{k}) (w - \hatw_{k}),  \\
L^\prox_{K-1}(w) &= \sum_{k=1}^{K-1} \hatL_{k}(\hatw_{k}) + (w - \hatw_{k})^\top\nabla \hatL_{k}(\hatw_{k}) + \frac{1}{2}(w - \hatw_{k})^\top \nabla^2 \hatL_{k}(\hatw_{k}) (w - \hatw_{k}).
\end{align*}
According to Lemma~\ref{lem:diff_function_val}, we know that
\begin{equation}\label{eq:gen_bound_1}
    F(w) \le \frac{1}{K}(L_K(w) + \tildeL_{K-1}(w)) + \frac{\rho}{2K}\sum_{k=1}^{K-1}\twonms{w - \hatw_k}^3.
\end{equation}
Note that here we choose $\delta=0$ in Lemma~\ref{lem:diff_function_val} since we do not consider the approximation error of the Hessian matrix. The next step is to use $\hatL_{K}(w) + L^\prox_{K-1}(w)$ to bound $L_K(w) + \tildeL_{K-1}(w)$. For this, we use the standard uniform convergence results with Rademacher complexity~\citep{mohri2018foundations}.

For any fixed $k\in[K]$, with probability at least $1-\delta$ over the random examples $(x_{k,i}, y_{k,i})$, $i=1,\ldots, n_k$, we have for all $w\in\W$
\begin{equation}\label{eq:rade_val}
  L_k(w) \le \hatL_k(w) + 2b\underbrace{\EE_\sigma\left[\frac{1}{n_k}\sup_{w\in\W}\sum_{i=1}^{n_k}\sigma_i \ell_k(w; x_{k,i}, y_{k,i})\right]}_{R^{(\ell)}_k} + 3b\sqrt{\frac{\log(2/\delta)}{n_k}},
\end{equation}
where $\sigma$ denotes a sequence of i.i.d. Rademacher random variables. Using normed Rademacher complexity, we can get with probability at least $1-\delta$,
\begin{equation}\label{eq:rade_grad}
    \sup_{w\in\W}\twonms{\nabla L_k(w) - \nabla \hatL_k(w)} \le  2g\underbrace{\EE_{\sigma}\left[ \frac{1}{n_k}\sup_{w\in\W}\twonm{\sum_{i=1}^{n_k} \sigma_i \nabla\ell_k(w;x_{k,i}, y_{k,i})}\right]}_{R^{(g)}_k} + 3g\sqrt{\frac{\log(4/\delta)}{2n_k}},
\end{equation}
and with probability at least $1-\delta$,
\begin{equation}\label{eq:rade_hess}
\begin{aligned}
    \sup_{w\in\W}\twonms{\nabla^2 L_k(w) - \nabla^2 \hatL_k(w)} \le  2h\underbrace{\EE_{\sigma}\left[\frac{1}{n_k} \sup_{w\in\W}\twonm{\sum_{i=1}^{n_k} \sigma_i \nabla^2 \ell_k(w;x_{k,i}, y_{k,i})}\right]}_{R^{(h)}_k}
    + 3h\sqrt{\frac{\log(4/\delta)}{2n_k}}.
\end{aligned}
\end{equation}
Then by
Cauchy–Schwarz inequality and union bound, we have with probability at least $1-\delta$,
\begin{align*}
    & L_K(w) + \tildeL_{K-1}(w) \\
   \le & \hatL_K(w) + L_{K-1}^\prox(w) + (L_K(w) - \hatL_K(w)) + \sum_{k=1}^{K-1} L_k(w) - \hatL_k(w) + \twonms{w-\hatw_k}\twonms{\nabla L_k(w) - \nabla \hatL_k(w)} \\
   &+ \frac{1}{2}\twonms{\nabla^2 L_k(w) - \nabla^2 \hatL_k(w)}\twonms{w - \hatw_k}^2 \\
   \le &  \hatL_K(w) + L_{K-1}^\prox(w) + 2bR^{(\ell)}_K + 3b\sqrt{\frac{\log(6K/\delta)}{n_K}} + \sum_{k=1}^{K-1} 2bR_k^{(\ell)} + 2g R_k^{(g)} + 2hR_k^{(h)} + 3(b+g+h)\sqrt{\frac{\log(12K/\delta)}{2n_k}}.
\end{align*}
The proof is then completed by combining this inequality with~\eqref{eq:gen_bound_1}.

\section{Experiment Details}\label{sec:experiment_detail}

We provide the hyper parameters that we used in all the experiments. As mentioned, for EWC, SI, and Kronecker algorithms, we introduce a regularization coefficient $\lambda$. In addition, for SI, there is another hyper parameter $\xi$ used in each update step (see Eq. (5) in Section 3 of~\cite{zenke2017continual} for the definition of $\xi$). The values of $\lambda$ and $\xi$ used in our experiments are provided in Table~\ref{tab:hyper}.

\begin{table}[ht]
\caption{Hyper parameters}\label{tab:hyper}
\centering
\setlength\tabcolsep{9pt}
\begin{tabular}{c|ccc}
\toprule
  &  Permuted MNIST &  Rotated MNIST & Split CIFAR	\\
\midrule
EWC $\lambda$ & 1.0 & 10.0 &  1.0 \\
SI $(\lambda, \xi)$ & (1.0, 1.0) & (1.0, 1.0) & (1.0, 0.1) \\
Kronecker $\lambda$ & 10.0 & 10.0 & 1.0\\
\bottomrule
\end{tabular}
\end{table}

\end{document}

%% file: commands.tex
\usepackage{amsthm}
\usepackage{setspace}
\usepackage{amsfonts}
\usepackage{bm}
\usepackage{xspace}
\usepackage{dsfont}
\usepackage{rotating}
\usepackage{graphicx} 
\usepackage{url}
\usepackage{xcolor,soul}
\usepackage{thm-restate}
\usepackage{standalone}
\usepackage{verbatim}
\usepackage{mdwmath}
\usepackage{amssymb}
\usepackage{amsmath}
\usepackage{xfrac}
\usepackage{mathtools}  
\usepackage{chngcntr}
\usepackage{tabulary}
\usepackage{booktabs}
\usepackage{epsfig}
\usepackage{wrapfig}
\usepackage{lipsum}
\usepackage{caption}
\usepackage{multirow}
\usepackage{footnote}
\usepackage{hyperref}
\usepackage{enumitem}
\usepackage{caption}
\usepackage{subcaption}

\newtheorem{theorem}{Theorem}
\newtheorem{definition}{Definition}

\newtheorem{lemma}{Lemma}

\newtheorem{prop}{Proposition}
\newtheorem{asm}{Assumption}

\newcommand{\ie}{\emph{i.e.,} }
\newcommand{\eg}{\emph{e.g.,} }

\newcommand{\R}{\mathbb{R}}

\newcommand{\gradF}{\nabla F}

\newcommand{\bigo}{\mathcal{O}}

\newcommand{\W}{\mathcal{W}}

\newcommand{\twonm}[1]{\left\|#1\right\|_2}
\newcommand{\twonms}[1]{\|#1\|_2}

\newcommand{\EE}{\ensuremath{\mathbb{E}}}

\newcommand{\innerps}[2]{\langle#1,#2\rangle}

\newcommand{\D}{\mathcal{D}}

\newcommand{\T}{\mathcal{T}}

\newcommand{\hatw}{\widehat{w}}

\newcommand{\X}{\mathcal{X}}
\newcommand{\Y}{\mathcal{Y}}

\newcommand{\hatL}{\widehat{L}}
\newcommand{\tildeL}{\widetilde{L}}
\newcommand{\tildeF}{\widetilde{F}}
\newcommand{\tildeD}{\widetilde{D}}

\newcommand{\prox}{\text{prox}}